\documentclass{llncs}
\usepackage{hyperref}
\usepackage{url}
\usepackage{amsmath}
\usepackage{comment}
\usepackage{algpseudocode}
\usepackage{algorithm}
\usepackage{graphicx}
\begin{document}

\def\xbar{\bar{x}}
\def\T{{\cal{T}}}
\def\grad{\nabla}
\def\vehicle{{\it Vehicle}~}
\def\ccat{{\it CCAT}~}
\def\covtype{{\it Covtype}~}
\def\mnist{{\it MNIST8m}~}

\title{A Distributed Algorithm for Training Nonlinear Kernel Machines}

\author{Dhruv Mahajan\inst{1} \and S. Sathiya Keerthi\inst{2} \and S. Sundararajan\inst{3}}
\institute{Microsoft Research \and CISL, Microsoft}

\maketitle

\begin{abstract}
This paper concerns the distributed training of nonlinear kernel machines on Map-Reduce. We show that a re-formulation of Nystr\"om approximation based solution which is solved using gradient based techniques is well suited for this, especially when it is necessary to work with a large number of basis points. The main advantages of this approach are: avoidance of computing the pseudo-inverse of the kernel sub-matrix corresponding to the basis points; simplicity and efficiency of the distributed part of the computations; and, friendliness to stage-wise addition of basis points. We implement the method using an AllReduce tree on Hadoop and demonstrate its value on a few large benchmark datasets.
\end{abstract}

\section{Introduction}
Recently, distributed machine learning has become an active area of research due to need for analysis on big data coming from various sources (for example, logs collected from a search engine). There has been a lot of focus on developing distributed learning algorithms for generalized linear 
models; see~\cite{bib1} and the references given there for key representative methods.

Training a nonlinear kernel machine is a lot harder due to the need for representing its solution as a linear combination of kernel functions involving the training examples. It involves either pre-computing and storing $n^2$ kernel entries or computing them on the fly in each iteration. Here, $n$ is the size of training data. In both the cases, doing $O(n^2)$ computation to get the exact solution is unavoidable. When the number of training examples goes beyond a million, it becomes necessary to resort to parallel/distributed solvers. Many methods have been suggested in the literature and most of them are given for SVMs. In subsection 1.1 we discuss some important ones.

A large body of literature~\cite{bib4,bib8,bib16,bib20,bib21,bib23,bib25,bib26,bib27,bib28} focuses on finding a low rank approximation of the kernel matrix. The approximate kernel is then used in place of the actual one to make training efficient and tractable. However, it is well known 
that for harder problems, the rank required to get reasonable performance can be quite large.

Much of the past work uses specialized hardware configurations (including clusters with MPI, multicore processors and GPUs) which are very communication efficient. There is little work~\cite{bib2} on scalable training of nonlinear kernel machines using Map-Reduce running on a cluster of commodity machines.

This paper concerns the efficient training of kernel machines (SVMs, Kernel logistic regression, Kernel ridge regression etc.) on large sized data. We propose a simple and effective approach for training the nonlinear kernel machines. We take the Nystr\"om route~\cite{bib21} of approximating the kernel matrix using a set of basis points that is much smaller than the training set, but we differ significantly from previous work in the way we train kernel machines using it. The key reason for this is the necessity of using a large number of basis points to reach a performance close to that of the true kernel machine. We use a formulation that avoids the computation of the pseudo-inverse of the kernel matrix involving the basis points. Another key idea that we use for achieving efficiency is the solution of the resulting optimization problem using a gradient-based method, in which the computation of the function, gradient and Hessian operations are carried out in a distributed fashion using Map-Reduce. Such a solution fits into the statistical query model outlined in~\cite{bib5}. Since iterative computation is expensive on Map-Reduce, we use an AllReduce tree of nodes~\cite{bib1} for our implementation. The computation that happens on AllReduce is simple as it involves only matrix-vector products. When the number of basis points is not very large, we choose them by distributed K-means clustering, and resort to random selection otherwise. (Effective selection of basis points is an area that needs further work.)

In summary, following are the key contributions of the paper.
\begin{enumerate}
\item We point out and reason that a re-formulation of the linearized kernel machine formulation based on Nystr\"om approximation is nicely suited for efficient solution. The formulation has two key advantages: the avoidance of pseudo-inverse computation; and, the ability to efficiently do stage-wise adaptation of basis elements.
\item We propose an AllReduce tree based implementation that is efficient and simple, with the distributed part consisting of only matrix-vector products.
\item We implement the proposed solution on Hadoop and demonstrate its value on some large benchmark datasets.
\end{enumerate}

\subsection{Related work}

There are methods~\cite{bib2,bib13} that partition the full data into several nodes, obtain a solution on each partition and combine them centrally. This ensemble solution is usually very sub-optimal. Some methods~\cite{bib6,bib7,bib11,bib19,bib22} do further processing of those solutions to come closer to full optimality, but this requires heavy communication of data, e.g., passing support vector data between nodes. There are methods that directly parallelize well-known sequential methods such as SMO and SVMlight~\cite{bib3,bib24}, but these involve communication costs over a large number of iterations.

P-packsvm~\cite{bib29} uses primal stochastic gradient descent (SGD) applied in the kernel feature space. Training examples are partitioned into various nodes. The most expensive computation of an SGD iteration is the calculation of the output associated with a given example. This is done by passing the example to all nodes, locally computing the sub-outputs associated with the nodes and then using MPI AllReduce operation to get the summed output to all nodes. A packing strategy is used to do $r$ iterations together; since $O(r^2)$ effort is involved, $r$ is kept small, say 100. P-packsvm is shown to work well on an MPI environment. On other distributed settings with communication overheads it can become slow due to the large number ($O(n)$) of iterations.

Hazan et al~\cite{bib12} obtain a decoupled solution on the partitioned data in each node. To reach optimality on the entire data, they use Fenchel duality ideas to set up an iterative loop where information is passed between nodes and the decoupled solutions are repeatedly solved. This method is similar in principle to the ADMM method~\cite{bib10}. For these methods, the number of iterations depends on the dataset. On hard datasets it can run into many thousands or more. Also, this number depends on the number of processors since reaching consensus between the decoupled solutions is harder with more processors.

There are methods that use a low rank approximation of the kernel. Chang et al~\cite{bib4} and Woodsend \& Gonzio~\cite{bib23} use low rank Cholesky factorization of the kernel matrix. There is a large literature based on using Nystr\"om method~\cite{bib8,bib16,bib20,bib21,bib25,bib26,bib27,bib28}. While the methods in this class are useful for getting decent sub-optimal performance, they are insufficient if we aim to achieve a performance matching the true kernel machine.

The paper is organized as follows. In section 2 we review methods of training kernel machines that are based on Nystr\"om approximation. In section 3 we give the details of our approach: the reasoning behind the choice of formulation, details of the distributed implementation, and methods of choosing basis points. Section 4 gives experiments on benchmark datasets that analyze our method and show its value. The appendix contains details of the datasets used in the paper.

\section{Methods based on reduced basis functions}

Let $\T=\{(x_i,y_i)\}_{i=1}^n$ denote the training set used for designing the kernel machine. We focus on binary classification and set $y_i\in\{1,-1\}$.
Let $k(x,\xbar)$ denote the kernel function. In this paper we will mainly use the Gaussian kernel, $k(x,\xbar)=\exp(-\|x-\xbar\|^2/2\sigma^2)$. Let $K\in R^{n\times n}$ be the kernel matrix with elements $K_{ij}=k(x_i,x_j)$. Training the kernel machine corresponds to solving the optimization problem,
\begin{equation}
\min \frac{\lambda}{2} \alpha^T K \alpha + L(K\alpha,y)
\label{kmin}
\end{equation}
where $L$ is the total loss function, $\sum_i l(o_i,y_i)$, $o_i$ is the $i$-th element of $K\alpha$, and $\lambda$ is the regularization constant.

\subsection{Nystr\"om method}

Nystr\"om method~\cite{bib21} gives a way of approximating the kernel matrix $K$. It selects a set of $m$ basis functions $(m<n)$, $\{\xbar_k\}_{k=1}^m$; this set is usually a subset of the training points, $\{x_i\}$. $K$ is approximated as
\begin{equation}
K \approx \tilde{K} = C W^+ C^T
\label{Ktil}
\end{equation}
where $C\in R^{n\times m}$, $C_{ik} = k(x_i,\xbar_k)$, $W\in R^{m\times m}$, $W_{kl}=k(\xbar_k,\xbar_l)$ and $W^+$ is the pseudo-inverse of $W$. For kernels such as the Gaussian kernel, if no two examples are identical then $W^+=W^{-1}$. $W^+$ is usually computed using the eigen-decomposition, $W=U\Lambda U^T$.

The complexity of computing $\tilde{K}$ is $O(nm^2+m^3)$ where the cost, $m^3$ comes from the need to compute the pseudo-inverse, $W^+$. When rank-$\tilde{m}$ approximation is used for $W$, the cost reduces to $O(nm\tilde{m} + \tilde{m}^3)$. Theoretical studies on Nystr\"om approximation based techniques have been done in the contexts of (a) reconstruction error~\cite{bib8}, and (b) objective function change and stability of solution (i.e., hypothesis learnt) obtained using kernel approximations with various kernel machines~\cite{bib9,bib20}. Of particular interest to this paper are the stability bounds available for support vector machines and kernel ridge regression in~\cite{bib20}, where the bounds are given in terms of the norm difference between $K$ and $\tilde{K}$.

\subsection{Linearized kernel machine}

Let $A=CU\Lambda^{-1/2}$ so that $\tilde{K} = AA^T$. Define $w=A^T\alpha$. If $K$ is replaced with $\tilde{K}$, then (\ref{kmin}) becomes a training problem for a linear kernel machine:
\begin{equation}
\min \frac{\lambda}{2} \|w\|^2 + L(Aw,y)
\label{Linear}
\end{equation}
This linearization approach was suggested by Zhang et al~\cite{bib28} and shown to be an effective way of training kernel machines via the Nystr\"om kernel approximation.

The following is worth pointing out and relevant when we discuss reduced basis functions. When the kernel machine is SVM, one should not confuse the basis points used in setting up (\ref{Linear}) (as well as(\ref{Ours}) below) with support vectors. Support vectors are training examples that get placed on or between the margin planes whereas basis points are used to represent the classifier function. The two coincide in the true SVM solution. In most datasets it is possible to approximate the true SVM classifier well using a number of basis points that is much smaller than the number of support vectors.

\section{Our approach}

An alternative to (\ref{Linear}) is to define $\beta=W^+C^T\alpha$ and write the equivalent formulation
\begin{equation}
\min f(\beta) = \frac{\lambda}{2} \beta^T W \beta + L(C\beta,y)
\label{Ours}
\end{equation}
Though (\ref{Ours}) is derived above via the Nystr\"om approximation of the kernel matrix, it can also be written down directly by approximating the true function space by the RKHS spanned by the functions defined by the basis points $\{\xbar_k\}_{k=1}^m$ and having $\beta^T W \beta$ as the norm.
Lin \& Lin~\cite{bib17} (see section VII of that paper) consider SVM loss functions and briefly discuss (\ref{Ours}).\footnote{Lee \& Mangasarian\cite{bib15} suggest replacing $\beta^T W\beta$ with $\|\beta^2\|$ so as to transform (\ref{Ours}) to the linear kernel machine form. But this takes one away from the true spirit of the nonlinear kernel machine, (\ref{kmin}).}

Although (\ref{Linear}) and (\ref{Ours}) are equivalent, Zhang et al~\cite{bib28} recommend using (\ref{Linear}) since it is in a linear kernel machine form; also, further computational efficiency can be brought in by avoiding a full eigen-decomposition of $W$ and instead doing only a low-rank decomposition~\cite{bib16}, i.e., $$W\approx\tilde{U}\tilde{\Lambda}\tilde{U}^T,$$ where $\tilde{U}\in R^{m\times\tilde{m}}$, $\tilde{\Lambda}\in R^{\tilde{m}\times\tilde{m}}$ and $\tilde{m}<m$. This leads to $$A\approx\tilde{A}=C\tilde{U}\tilde{\Lambda}^{-1/2}.$$ All these are fine if efficiency is of paramount importance and sub-optimal generalization performance can be tolerated. It turns out that, in many large scale datasets, the $m$ and $\tilde{m}$ values needed for achieving performance close to that of the true kernel machine (\ref{kmin}) is usually a sizeable fraction of $n$, say $m, \tilde{m} \approx 0.1n$. Thus, computing the eigen-decomposition of $W$ [or an approximate decomposition of dimension $\tilde{m}$] is expensive, requiring $O(m^3)$ [$O(m\tilde{m}^2)$] effort. Furthermore, if the transformed data matrix $A$ needs to be formed so as to solve (\ref{Linear}) using a standard linear kernel machine solver, it requires $O(nm^2)$ [$O(nm\tilde{m})$] effort. Thus the solution of (\ref{Linear}) can be quite expensive. Both the above mentioned expensive computations are avoided in (\ref{Ours}).

Not only does (\ref{Ours}) avoid expensive computations related to the eigen-decomposition of $W$, but it is also well suited to efficient distributed solution. Depending on the type of kernel machine and loss function, a suitable method can be used to solve (\ref{Ours}). We can use a gradient based technique~\cite{bib18} if the objective function is differentiable, and a non-differentiable solver such as the bundle method otherwise. Let us consider binary classification and the differentiable loss, $$l=\frac{1}{2} \max(1-y_io_i,0)^2,$$ where $o_i=c_i\beta$ and $c_i$ is the $i$-th row of $C$. In such a case, $$\grad_\beta f = \lambda W\beta + C^TD(C\beta-y),$$ where $D$ is a diagonal matrix such that 
\begin{equation}
\nonumber
    D_{ii}= 
\begin{cases}
    1,& \text{if } 1-y_io_i>0\\
    0,              & \text{otherwise}
\end{cases}
\end{equation}
If $\grad_\beta f$ is computed as $\lambda (W\beta) + C^T(D(C\beta-y))$, the cost is $O(nm)$. A suitable gradient based technique such as TRON (Trust Region Newton)~\cite{bib18} can be used to minimize $f$ in (\ref{Ours}).TRON also requires an efficient method for computing $Hd$ where $H$ is the (pseudo) Hessian of $f$ and $d\in R^m$. For the $f$ in (\ref{Ours}), $$Hd=(\lambda W + C^TDC)d;$$ like the gradient computation it can also be computed in $O(nm)$ time. Typically, TRON requires at most a few hundred iterations, with each iteration involving one function/gradient computation and a few $Hd$ computations. Thus, empirically, the solution of (\ref{Ours}) can be solved in $O(nm)$ time.

The expensiveness of (\ref{Linear}) as $m$ grows can be demonstrated even on a single machine. Table~\ref{vehicle-table} compares the costs associated with setting up and solving the formulations (\ref{Ours}) and (\ref{Linear}) on the \vehicle dataset~\cite{bib14} for the hyperparameter setting, $\lambda=8.0$ and $\sigma=2.0$. Clearly, the cost associated with the computation of $A$ (computation of $W^+$ is a key part of this) becomes very large when $m$ grows large ($O(m^3)$ growth). On the other hand, the total time for solving (\ref{Ours}) only grows linearly with $m$.

\begin{table}[t]
\caption{Comparing formulations (\ref{Ours}) and (\ref{Linear}) on the \vehicle dataset}
\label{vehicle-table}
\begin{center}
\begin{tabular}{|l|l|c|c|c|c|}
\hline
\multicolumn{2}{|c|}{$m$ $\rightarrow$} &   100   &  1000  &  10000  \\
\hline
Formulation (\ref{Ours})   & Total time (secs)    &  $87.4$    &  $693$    &  $6704$     \\
\hline
Formulation (\ref{Linear}) & Total time (secs)    & $88.3$   & $713$ & $9340$ \\
        & Fraction of time for $A$    & 0.0017  & 0.0148 & 0.2893  \\
\hline
\end{tabular}
\end{center}
\end{table}

\vspace*{0.1in}
{\noindent{{\bf Stage-wise addition of basis points.}}}
Suppose we want to add basis functions in a graded fashion, i.e., increase $m$ in stages. For such a mode of operation, (\ref{Linear}) requires incremental computation of the SVD of $W$, which is messy and expensive. On the other hand, solution of (\ref{Ours}) does not pose any issues. In fact, one can use the $\beta$ obtained for a set of basis points to initialize a good $\beta$ when new basis points are added: we just set the $\beta$ variables corresponding to the newly added basis points to zero. Note also that only the part of the kernel matrix corresponding to the newly added basis points needs to be computed. All these add to the advantages associated with (\ref{Ours}).

Since (\ref{Ours}) is directly associated with the Nystr\"om method, results for quality of approximation mentioned in subsection 2.1 will hold for its solution too. Though the formulations (\ref{Linear}) and (\ref{Ours}) involve different (but related) variables, the error bounds in~\cite{bib20} for (\ref{Linear}) can be easily modified for (\ref{Ours}).

\begin{algorithm*}[t]
\caption{Map-Reduce TRON Algorithm for (\ref{Ours})}
\begin{algorithmic}
\vspace*{0.08in}
\State 1. {\bf Data loading.} The $n$ training examples are randomly distributed on the $p$ nodes.\vspace*{0.08in}
\State 2. {\bf Communication of basis points.} Each node chooses $\frac{m}{p}$ basis points randomly from its data. The basis points are broadcast to all nodes.\vspace*{0.08in}
\State 3. {\bf Kernel computation.} Each node constructs its row block of the kernel matrix $C$. Note that the corresponding row block of $W$ is a subset of the $C$ row block and so it doesn't have to be computed.\vspace*{0.08in}
\State 4. {\bf TRON optimization.} One of the $p$ nodes is identified as the master and it carries out the TRON iterations. The following steps sketch how the function, gradient and $Hd$ operations are carried out.\vspace*{0.08in}
\begin{itemize}
\State 4a. {\bf Function computation.} $\beta$ is broadcast to all nodes. $\beta^TW\beta$ is computed as $$\sum_{j=1}^p (\beta)_j^T (W\beta)_j,$$ where $(z)_j$ denotes the $j$-th row block of $z$, the nodes compute the elements of the summation in parallel, and the summation is carried out using AllReduce. For the second term in (\ref{Ours}) we compute the various row blocks of $C\beta$ and the corresponding pieces needed for $L(C\beta,y)$ in parallel and compute it using AllReduce.\vspace*{0.08in}
\State 4b. {\bf Gradient computation.} Compute the gradient as $$\sum_j \lambda (W\beta)_j + (C)_j^T((D)_j(C\beta-y)_j),$$ doing the inner elements in parallel and summing using AllReduce.\vspace*{0.08in}
\State 4c. {\bf $\mathbf Hd$ computation.} The sequence of computations is same as the gradient computation with $\beta$ replaced by $d$ and $y$ set to $0$.
\end{itemize}
\end{algorithmic}
\end{algorithm*}

\subsection{Distributed solution}
\label{distsol}

On large scale problems, the function, gradient and $Hd$ computations form the main bulk of the cost of minimizing $f$ using TRON. (All other computations in TRON require only $O(m)$ effort.) The attractive part is that these computations are based on matrix vector multiplications, which is well suited for distributed computation.

We implement our method using Map-Reduce on Hadoop. It is known that this system is ill-suited for iterative algorithms such as TRON. To overcome this problem, like~\cite{bib1} we employ a set of $p$ nodes arranged in an AllReduce tree arrangement. This set-up will be used for all broadcast, gather and reduce operations.

Our current implementation uses up to 200 nodes ($p$=200). Since the number of nodes is not large, we simply go for a row partitioning of $W$ and $C$ over the nodes. (If a large set of nodes is available then it is better to have hyper-nodes handle row partitioning and attach an AllReduce tree of nodes to each hyper-node to handle column level partitioning. Such an arrangement will reduce latencies in the system.)
Algorithm 1 gives the implementation details for the case where the set of $m$ basis points, $\{\xbar_k\}$ are chosen randomly from the training examples. In section 4 we will discuss other ways of choosing basis points.

If the number of TRON iterations is taken as a constant, Algorithm 1 has time complexity $O(nm/p)$. The communication complexity is $O(m^2/p)$, dictated by step 2. Each node requires a memory of $O(nm/p)$. If such a memory allocation is beyond the capacity of one node, then kernel caching ideas (similar to those used in standard solvers such as SMO and SMOlight) that keep frequently used kernel elements in the available memory cache and compute other kernel elements on the fly, need to be employed.

The communication cost associated with TRON optimization (step 4 of Algorithm 1) is $O(n)$ for steps 4a and 4b (function and gradient), and $O(m)$ for step 4c ($Hd$), which are not high. However, these steps are called many times: the number of calls to 4a and 4b is equal to $N$, the number of TRON iterations, which is typically around 300; the number of calls to step 4c is around 3N, which is around 900. Thus, communication-wise, step 4 is efficient. However, if the implementation of AllReduce is crude and such that there is a high latency each time a communication is made, then the latencies accumulated over 1500 calls of 4a, 4b or 4c can be substantial. Unfortunately our current implementation suffers from such a latency; we will point out the ill-effects of this in section 4.

On a related note, we want to mention that, P-packsvm~\cite{bib29}, which is one of the best parallel nonlinear kernel machine solvers reported in the literature, communicates a small pack of information in each of its iterations but the number of iterations is very large: $O(n)$. Fortunately, P-packsvm is run on a professional MPI cluster with negligible latency. Thus, that implementation does not suffer.

\subsection{Selecting Basis functions}

It has been observed in the Nystr\"om approximation literature~\cite{bib26} that selection of basis functions is important for achieving top performance when keeping $m$ to be a small fraction of $n$. Gradient based tuning of the basis function parameters to optimize $f$ is possible~\cite{bib14}, but it is sequential and quite involved. Cluster centers obtained via K-means clustering form good basis functions when Gaussian kernel is used~\cite{bib26}. However it becomes substantially expensive\footnote{The cost is nearly $N_{K-means}$ times the cost of computing the kernel sub-matrix $C$, where $N_{K-means}$ is the number of iterations in the K-means algorithm.} when the number of basis points, $m$ is large. Moreover, since the basis points do not form a subset of the training points, $W$ needs to be computed. We have found that using an expensive method such as K-means clustering is beneficial when $m$ is small. Table~\ref{kmeans-table} gives results of an experiment on the \covtype dataset. The K-means clustering method was made to use only three iterations. At large $m$, K-means clustering forms a significant fraction of the total time, but doesn't contribute much to improving accuracy. Thus, we use a (distributed) K-means algorithm when $m$ is not too large, and switch to random selection of basis points otherwise.\footnote{We also do not use K-means clustering when the number of features in the dataset is large, e.g., \ccat (see section 4).} Note that, with random selection, it is possible to judiciously choose the basis points using a data-dependent distribution~\cite{bib8}.

\begin{table}[t]
\caption{Comparing K-means and Random basis selection on \covtype dataset. All times are in seconds.}
\label{kmeans-table}
\begin{center}
\begin{tabular}{|c|c|c|c|c|c|c|}
\hline
   & \multicolumn{3}{|c|}{$m=1600$} & \multicolumn{3}{|c|}{$m=51200$} \\
\hline
   & {Test set} & {K-means} & {Total} & {Test set} & {K-means} & {Total} \\
   & {Accuracy} & {Time}    & {Time}  & {Accuracy} & {Time}    & {Time} \\
\hline
{K-means}        & 0.8087  & 49.49 & 355.97  & 0.9493  & 1399.28  &  3899.97 \\
\hline
{Random Basis}   & 0.7932  &  -    & 300.98  & 0.9428  & -        &  2678.74 \\
\hline
\end{tabular}
\end{center}
\end{table}

\section{Experiments}

In this section we describe some experiments using our implementation on some benchmark datasets. Algorithm 1 was implemented in a Hadoop cluster with 200 nodes. Each node has a 2.3 GHz CPU with a memory of 16 GB. The AllReduce tree is a natively built code in this set-up. Thus our current implementation is crude, and has associated issues such as latency in communication. We say more about the ill-effects of this below. With effort, a lot better AllReduce tree implementation is easily possible.

\subsection{Datasets used}

We used four benchmark datasets~\cite{bib14,bib29}. Their parameters are given in Table~\ref{tab:data}. The parameters are as follows: $n$ and $n_{test}$ are the numbers of training and test examples; $d$ is the number of features; the hyper-parameters are $\lambda$, the regularization constant in (\ref{Ours}) and $\sigma$ is the kernel-width parameter in the Gaussian kernel function, $k(x,\xbar)=\exp(-\|x-\xbar\|^2/2\sigma^2)$.

\begin{table}
\caption{Datasets}
\label{tab:data}
\begin{center}
\begin{tabular}{|c|c|c|c|c|c|c|}
\hline
    & $n$ & $n_{test}$ & $d$ & $\lambda$ & $\sigma$\\
\hline
\vehicle  & 78,823 & 19,705 & 100 & 8 & 2 \\
\hline
\covtype & 522,910 & 58,102 & 54 & 0.005 & 0.09\\
\hline
\ccat & 781,265 & 23,149 & 47,236 & 8 & 0.7\\
\hline
\mnist & 8,000,000 & 10,000 & 784 & 8 & 7\\
\hline
\end{tabular}
\end{center}
\end{table}

We believe that these datasets are sufficient to make important observations due to following reasons: (1) our method targets datasets that require large $m$, and, these datasets require large $m$ (see accuracy vs $m$ plots in Figure~1),  (2) we want to study the effect of number of features in the overall cost (see the wide variation in our datasets), as they influence kernel computation time, and, (3) large number of examples (n) – e.g., \mnist has $8$ Million examples.

\begin{figure*}[t]
\label{mplot}
\begin{center}
\vspace*{-0.8in}
   \includegraphics[width=0.49\linewidth]{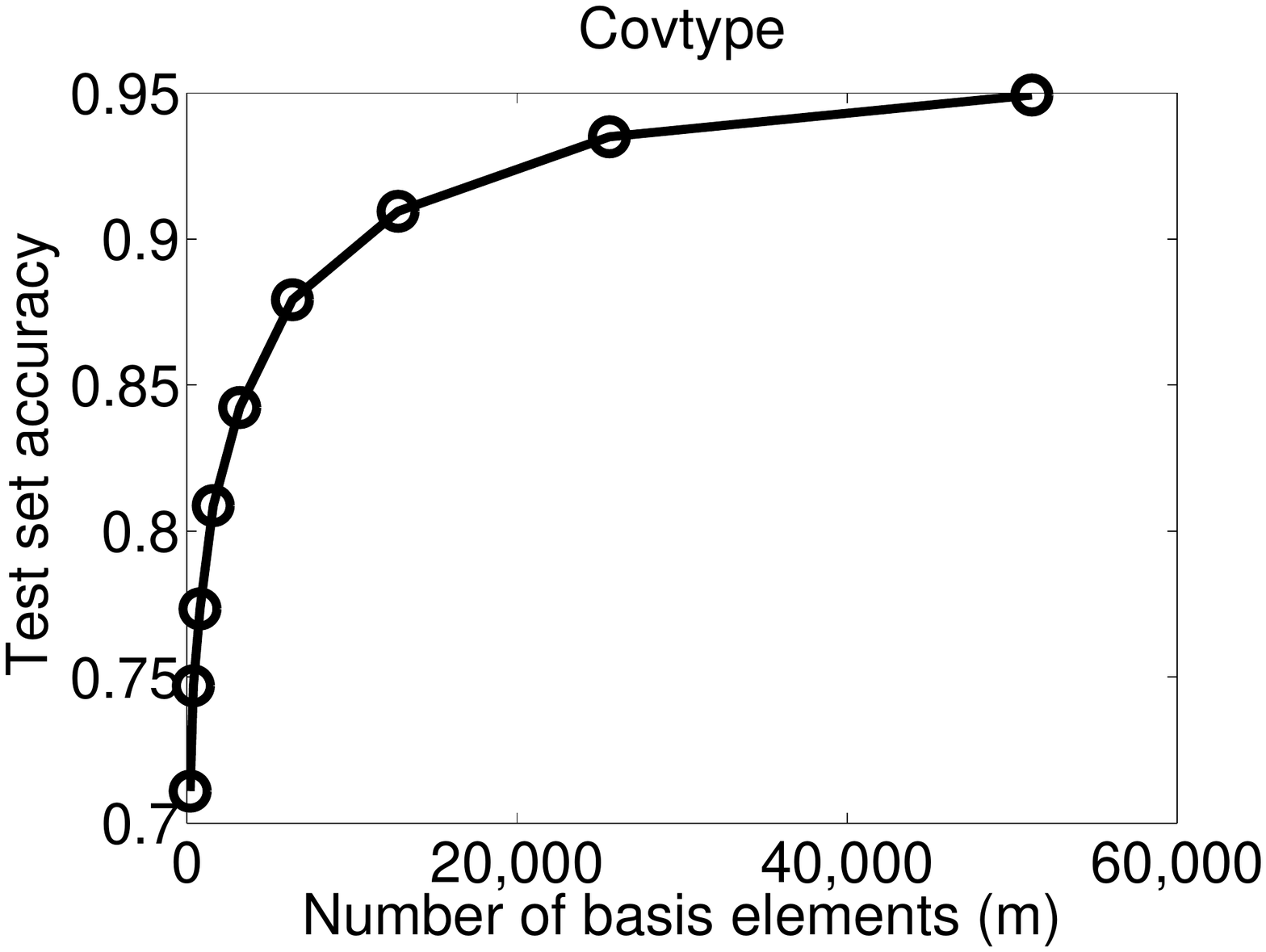}
   \includegraphics[width=0.49\linewidth]{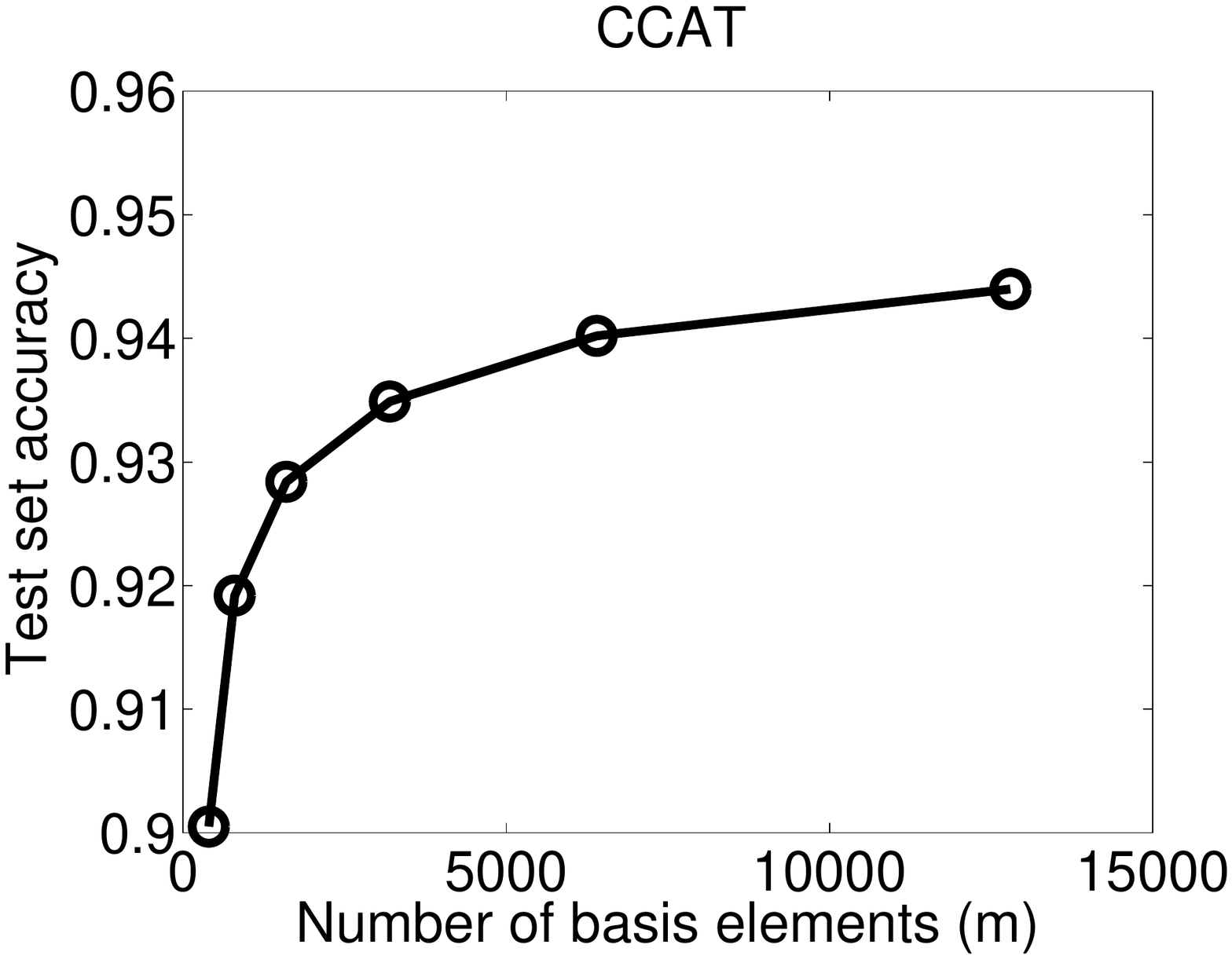}
\end{center}
\vspace*{-0.7in}
\caption{Test set performance versus $m$ for \covtype (left) and \ccat (right)}
\end{figure*}

\subsection{Need for large $\mathbf m$}

Most large scale datasets are such that the learning curve stabilizes only at a large number of examples. For these, a large number of basis points (many tens of thousands or more) are needed to reach a performance close to that of the true kernel machine. Figure~1 gives plots of test set accuracy versus $m$ for \covtype and \ccat. As expected, for small $m$ the accuracy increases very fast, while the gains are much less on going to higher $m$ values. For \covtype it is well known that the number of support vectors is more than half of the training set. Hence, the curve does not stabilize even for $m=51200$. Moreover, as discussed earlier in Section~3, our method gives the additional flexibility to incrementally add basis points and stop at the desired point on the performance curve. It is worth noting that for a large $m$ such as $m=51200$, the linearized kernel machine approach [28] becomes extremely expensive.

\subsection{Slicing of computational costs}

\begin{table}[t]
\caption{Costs of different steps of Algorithm 1. All times are in seconds.}
\label{slice-table}
\begin{center}
\begin{tabular}{|c|c|c|c|c|}
\hline
  $m$ & \multicolumn{4}{|c|}{Steps of Algorithm 1} \\
	\hline
	\multicolumn{5}{|c|}{\covtype, nodes - $200$} \\
\hline
        &    1    &    2     &    3      &    4       \\
\hline
200     & 0.24    & 1.36     & 2.15      & 160.92     \\
\hline
3200    & 0.22    & 1.50     & 34.22     & 480.06     \\
\hline
51200   & 0.23    & 3.50     & 627.26    & 1869.70    \\
\hline
\multicolumn{5}{|c|}{\mnist, nodes - $200$} \\
\hline
1000     & 228    & 18.99    & 868.655      & 89.21     \\
\hline
10000    & 229    & 80.36     & 7928.40     & 850.69     \\
\hline
\multicolumn{5}{|c|}{\ccat, nodes - $200$} \\
\hline
400     & 14.82    & 8.77     & 222.67      & 329.81     \\
\hline
3200    & 14.51   & 21.22     & 1451.19    & 1395.34    \\
\hline
12800   & 15.21    & 68.68     & 5817.26    & 3648.19    \\
\hline
\multicolumn{5}{|c|}{\vehicle, nodes - $1$} \\
\hline
100     & 20.42    & 51.45     & 596.47      & 71.25     \\
\hline
1000    & 20.00   & 11.04     & 6029.32    & 694.57    \\
\hline
10000   & 19.92    & 6.56     & 61778.90    &  5000.00    \\
\hline
\end{tabular}
\end{center}
\label{tab:algosteps}
\end{table}

Table~\ref{tab:algosteps} gives an idea of the times associated with various steps of Algorithm 1, for different datasets. The data loading time is pretty much a constant and small. Communication of basis points is done once. For $m$ basis vectors we need to transmit $mk$ non-zeros, where $k$ is number of non-zeros in each row. But kernel computation cost (step 3) per node is $O(nmk/P)$, which is typically much more than the basis broadcast cost (step 2). The $TRON$ algorithm cost is $O(nm/P)$ with several iterations. Hence, the basis broadcast cost is not significant.

Whether TRON or Kernel computation is the bottleneck depends on the interplay between dimensionality ($d$) and sparsity ($k$), number of examples ($n$), iterations required and regularizer ($\lambda$). For higher dimensional datasets like \mnist ($d = 784$), kernel computation is dominant (step 3). On the other hand, \covtype requires several hundred of $TRON$ iterations. Hence, $TRON$ optimization dominates the cost (step 4). Similarly, if regualrizer $\lambda$ is small, iterations start increasing and we expect the $TRON$ cost to start dominating. Nevertheless, both steps are parallelizable and hence we do not expect this factor to play a major role.

\begin{figure*}[t]
\label{speedupplot}
\begin{center}
\vspace*{-0.8in}
   \includegraphics[width=0.49\linewidth]{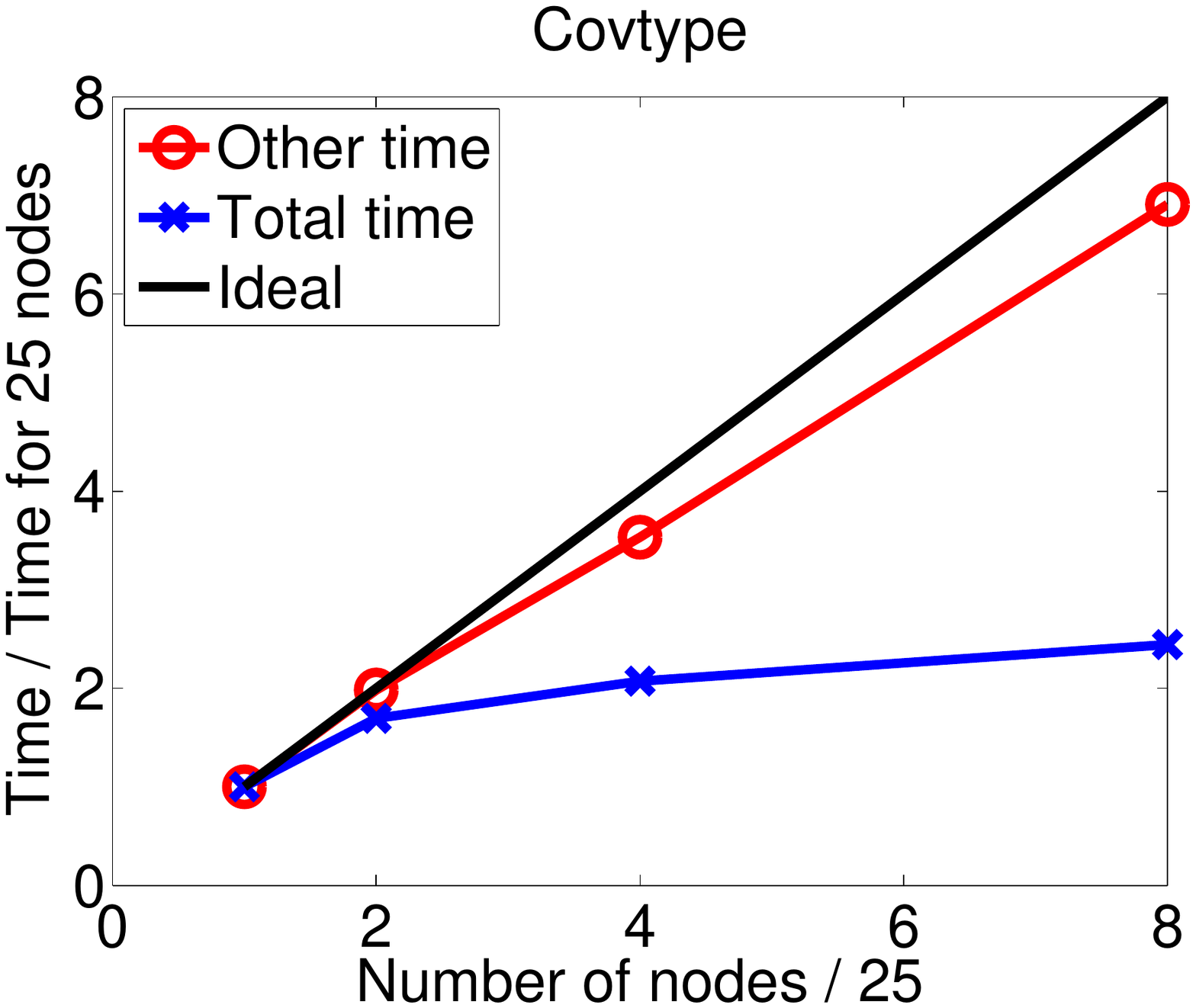}
   \includegraphics[width=0.49\linewidth]{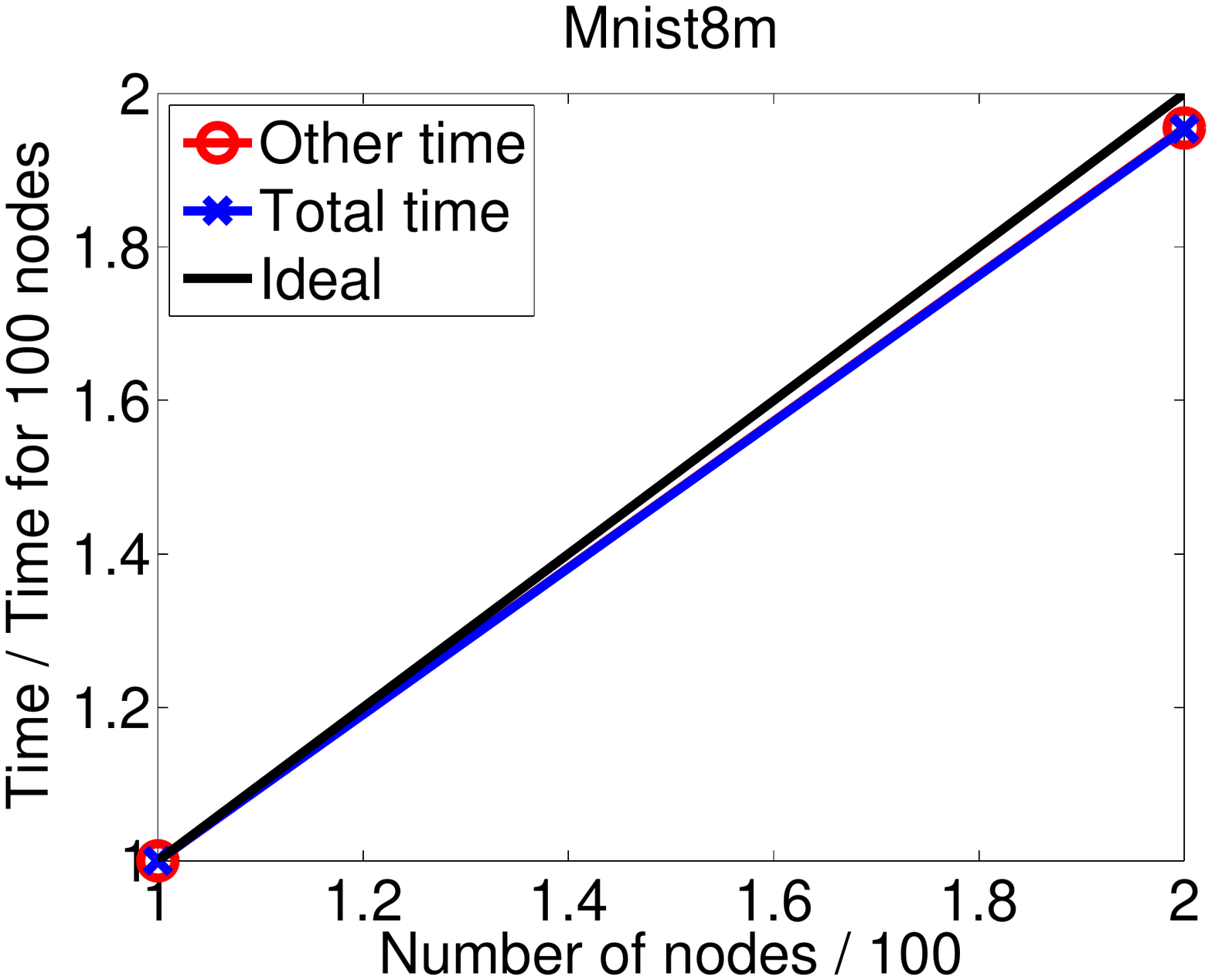} 
\end{center}
\vspace*{-0.7in}
\caption{Speed-up plots for \covtype (left) and \mnist (right). For \covtype, 25 nodes is used as reference and for \mnist, 100 nodes is used as reference. {\it Other time} refers to the time associated with all steps of Algorithm 1 except the TRON optimization step (step 4).}
\end{figure*}

\subsection{Parallel speed-up}

Let us now analyze the speed-up that can be achieved with many nodes. Figure~2 gives speed-up data for \covtype and \mnist. At the end of subsection~\ref{distsol} we discussed issues related to communication latency. Since communication latencies related to step 4 play an important role, let us discuss this in some detail. The number of calls to steps 4a, 4b or 4c is around $5N$ where $N$ is the number of TRON iterations. Let us write one instance communication cost in the form $C+DB$ where $C$ is communication latency, $D$ is the cost of communication per byte after leaving out latency, and $B$ is the number of bytes transferred. Over $5N$ calls the total cost related to communication in step 4 is $5N(C+DB)$. In step 4, $B$ is small. Also, with our current implementation of AllReduce being crude, the term $5NC$ dominates. If the local computation time is comparatively small, this term becomes bothersome. In \covtype, $N$ is a few thousands, local computation time is small, and the $5NC$ term forms a major fraction of the total time. Unfortunately, this term is an overhead that is independent of the number of nodes, and so speed-up suffers: see the speed-up plot for {\it Total time} for \covtype in Figure~2. Note that, in the same plot, the speed-up curve for {\it Other time}, which is the time for all steps other than TRON, is very good. Interestingly, for \mnist, the local computation time is heavier than communication latencies and communication time, and so the speed-up on {\it Total time} itself is close to linear.

\subsection{Comparison with P-packsvm.}

Any method that requires SVD is infeasible for large $m$; also, as we look at various computational costs, it is clear theoretically that any SVD based method is just not comparable in terms of speed with our method. Due to these reasons, we did not include any such method for comparison. For test set performance, it is clear that for a given $m$ basis vectors, our method will give same accuracy as any other linearized kernel machine using the same kernel matrix (defined using those basis vectors) and they differ only in parameterization; so, we did not include them. On the other hand, it is important to compare with a full kernel method which does not make any kernel approximation. P-packsvm~\cite{bib29} falls under this category.

P-packsvm~\cite{bib29} is one of the fastest known parallel solvers for nonlinear kernel machines. Though we have not implemented it ourselves,\footnote{If P-packsvm was implemented on our native Hadoop AllReduce system, it will be hugely inefficient due to the accumulation of communication latencies over a large number ($O(n)$) of iterations.} it is useful to make a rough comparison on the largest dataset, \mnist with $8$ million examples. The nodes used in the two works are comparable in speed and memory. Table~\ref{tab:ppackcomp} shows the comparison between our method ($m=10,000$), and P-packsvm run for $1$ epoch only. Note that our method significantly outperforms P-packsvm in terms of time taken, while it is also slightly better in terms of accuracy. The results are particularly impressive given P-packsvm uses an MPI cluster of 512 nodes, whereas we employ a native AllReduce implementation on Hadoop, with 200 nodes. Note that although it is possible to get better accuracies by running P-packsvm with multiple epochs, this will incur a significantly higher cost.

\begin{table}
\caption{Comparing P-packsvm and our method on \mnist dataset.}
\label{tab:ppackcomp}
\begin{center}
\begin{tabular}{|c|c|c|c|c|c|c|}
\hline
   & {Number of } & {Test set} & {Total Time}\\
   & {Nodes} & {Accuracy}    & {(Secs.)}  \\
\hline
{P-packsvm}        & 512 & 0.9948 & 12880 \\
\hline
{Our Method}   & 200  &  0.9963  & 8779  \\
\hline
\end{tabular}
\end{center}
\label{tab:ppackcomp}
\vspace*{-0.1in}
\end{table}

\section{Discussion}
In this section, we will discuss some of the future directions for the efficient distributed training of nonlinear kernel machines.

The proposed and related approaches basically operate in two steps: a) approximate the kernel matrix, and b) use this approximate kernel in training. The two step approach has the following disadvantage. Given a fixed number of basis vectors, the optimal basis obtained for kernel approximation in the Frobenious norm sense may not be good in achieving good generalization performance for the task (classification, regression etc.) at hand. Due to the sub-optimal nature of the approach, the number of basis vectors required to achieve a given generalization performance can be large. On the other hand, joint optimization of the basis vectors and model parameters has the potential of reducing the number of basis significantly. However, the resulting problem is non-convex, and designing a suitable regularizer on the basis to reach a good solution is an interesting future work. Recently, random Fourier basis~\cite{bib30} have become popular since they provide an unbiased estimate for kernel matrix. One can also explore the possibility 
of replacing random sampling by data dependent sampling geared towards the task at hand.

Alternatively, one can try to solve the original objective function without any approximation. Recently, Pechyony et al.~\cite{bib31} proposed a parallel block decomposition method to solve the linear SVM in the dual domain. It is an iterative method where in every iteration, each node optimizes over a subset of variables, while keeping variables from the other nodes fixed; the local solutions are then combined together to get the overall update. This method can be extended to non-linear kernels also, with suitable modifications to the subproblems in each node, and the combination step. Proving the global convergence of such an algorithm is an interesting future work. In case the raw data matrix fits in the memory of a single machine, each node can possibly choose points randomly (with overlap) in every iteration to improve the rate of convergence. Other heuristic ways of communicating some important points across machines can also be explored.

\section{Conclusion}

In this paper we have proposed a method for distributed training of nonlinear kernel machines that is efficient and simple. With upcoming Hadoop Map-Reduce systems having very good support for iterative computation and AllReduce, our approach can be very useful. A key future direction is to look at efficient distributed ways of tuning basis points to directly optimize the training objective function.

\bibliographystyle{plain}
\bibliography{kernel}

\end{document}